\documentclass[runningheads]{llncs}
\usepackage{graphicx}
\usepackage{subfigure}
\usepackage{cite} 

\usepackage{amsmath,amssymb} 
\usepackage{color}

\begin{document}
\title{Convolutional Neural Network\\for Trajectory Prediction} 

%
\author{Nishant Nikhil\inst{1,2} \and
Brendan Tran Morris\inst{2}\orcidID{0000-0002-8592-8806}}
%
\authorrunning{N. Nikhil and B. T. Morris}
%

\institute{Indian Institute of Technology Kharagpur\\
\email{nishantnikhil@iitkgp.ac.in} \\
\and
University of Nevada, Las Vegas\\
\email{brendan.morris@unlv.edu}
}

\maketitle              

\begin{abstract}
Predicting trajectories of pedestrians is quintessential for autonomous robots which share the same environment with humans. In order to effectively and safely interact with humans, trajectory prediction needs to be both precise and computationally efficient.  
In this work, we propose a convolutional neural network (CNN) based human trajectory prediction approach. Unlike more recent LSTM-based moles which attend sequentially to each frame, our model supports increased parallelism and effective temporal representation. The proposed compact CNN model is faster than the current approaches yet still yields competitive results.
\keywords{Convolutional Neural Network, Trajectory Prediction, Anticipating Human Behavior}

\end{abstract}

\section{Introduction}

Autonomous robots like self-driving cars on a road or a food-delivery robot in a restaurant must share the same space with humans. In order to do so in a safe and acceptable manner, these robots must be able to understand and cooperate with humans.  One task of paramount importance for avoiding collisions and for smooth maneuvering is to accurately predict the future trajectories of humans in their shared space.  Further, given the wide diversity of platforms and environments for which prediction may be required (e.g. small robots with limited computing capabilities or without connectivity to cloud computing resources), simple models with better time complexity are desired.  


Traditionally, hand-crafted features were used for trajectory prediction and modeling motion of pedestrians' trajectory with respect to others surrounding them. \cite{Antonini:77712} propose a discrete choice framework for pedestrian dynamics, modeling short-term behavior of individuals as a response to the presence of other pedestrians. The Social Force model \cite{sforce} incorporates two interactive forces for microsimulation of crowds.  
Attractive forces guiding the pedestrians towards their goal and repulsive forces for encouraging collision avoidance in-between the pedestrians and in-between a pedestrian and environmental obstacles. Yamaguchi et al. \cite{whoareyouwith} solves the same problem as an energy minimization problem.  While successful, hand-crafted features are hard to scale since influencing factors must be described explicitly.  

In recent years, Deep Neural Networks (DNN) have been utilized for the trajectory prediction task since they utilize a data-driven approach to tease out relationships and influences which may not have been apparent.  These DNN-based approaches \cite{gupta2018social, Alahi_2016_CVPR, sadeghian2018sophie, DBLP:journals/corr/FernandoDSF17} have demonstrated impressive results. Almost all of these approaches are based on Recurrent Neural Networks (RNNs) \cite{mikolov2010recurrent} since a trajectory is a temporal sequence. As RNNs share parameters across time, they are capable of conditioning the model on all previous positions of a trajectory. Although theoretically, RNNs can retain information from all previous words of a sentence, practically they fail at handling long-term dependencies. Also, RNNs are prone to the vanishing and exploding gradient problems when dealing with long sequences. 

Long Short-Term Memory (LSTM) networks \cite{Hochreiter:1997:LSM:1246443.1246450}, a special kind of RNN architecture, were designed to address these problems. Although LSTMs have been found to address the sequence based problems effectively but they need quite a bit of task-specific engineering like clipping gradients. Also in RNNs, predictions for later time-steps must wait for the predictions from preceding time-steps and hence can't be parallelized during training or inference time.  

Recently, Convolutional Neural Network (CNN) based architectures have provided encouraging results in sequence-to-sequence tasks \cite{DBLP:journals/corr/abs-1803-01271} like machine translation \cite{DBLP:journals/corr/GehringAGYD17, DBLP:journals/corr/VaswaniSPUJGKP17}, image generation \cite{Oord:2016:CIG:3157382.3157633} and Image Captioning \cite{AnejaConvImgCap17}. Inspired by these, we study CNNs for the task of trajectory prediction. This is the first work we are aware of to use an end-to-end convolutional architecture for trajectory prediction (Deo and Trivedi \cite{Deo2018ConvolutionalSP} used convolutional pooling for incorporating social context from hidden states of the LSTM network). We believe the CNN is superior to LSTM for temporal modeling since trajectories are continuous in nature, do not have complicated ``state'', and have high spatial and temporal correlation which can be exploited by computationally efficient convolution operations.  

The major contribution of the work can be summarized as proposing a fast CNN-based model for trajectory prediction that is competitive with more complicated state-of-the-art LSTM-based techniques which require more contextual information. We discuss our CNN architecture in Section 2. Section 3 provides an experimental evaluation to highlight the efficacy of our approach and value due to simplicity. Finally, in Section 4, we conclude the paper with closing remarks.

\section{Trajectory Prediction Method}

Recent work in prediction has utilized recurrent networks to model temporal dependencies and sequence-like nature of trajectories using LSTMs.  Most efforts in this area look to augment position input with social \cite{Alahi_2016_CVPR, gupta2018social} or scene \cite{sadeghian2018sophie, 8354239} context resulting in more complicated architectures.  In contrast, our work seeks to simplify the network architecture and make more direct use of trajectory structure (spatio-temporal consistency) by using highly efficient convolutions temporal support. 

\subsection{Problem Setup}

For trajectory prediction, we are given the trajectory of all the pedestrians. It is assumed that each scene is pre-processed and we have the spatial co-ordinates of every $i$-th pedestrian at time $t$ as $X_t = (x_t^i, y_t^i)$. That is, we have the pedestrian trajectory data as $X = \{X_1, X_2, X_3, X_4 ,\ldots, X_n\}$ for time steps $t=1,2,\ldots, T_{obs}$. Note: for simiplicity the pedestrian superscript $i$ is not listed. We have to predict the future trajectories of all the pedestrians for time steps $t= T_{obs+1},\ldots,T_{pred}$ as $\hat{Y} = \{\hat{Y}_1, \hat{Y}_2, \hat{Y}_3, \hat{Y}_4, \ldots, \hat{Y}_m\}$ all at once.

\subsection{LSTM-Based Frameworks}
Most current reserach in trajectory prediction has utilized LSTM cells for handling temporal dependencies. The working of LSTM cells are governed by the following equations:
\begin{equation}
f_t = \sigma_g(W_{f} X_t + U_{f} h_{t-1} + b_f) \\
\end{equation}
\begin{equation}
i_t = \sigma_g(W_{i} X_t + U_{i} h_{t-1} + b_i) \\
\end{equation}
\begin{equation}
o_t = \sigma_g(W_{o} X_t + U_{o} h_{t-1} + b_o) \\
\end{equation}
\begin{equation}
c_t = f_t \circ c_{t-1} + i_t \circ \sigma_c(W_{c} x_t + U_{c} h_{t-1} + b_c) \\
\end{equation}
\begin{equation}
h_t = o_t \circ \sigma_h(c_t)
\end{equation}

In these equations, $X_t$ is the input vector to the LSTM unit, $f_t$ is the forget gate's activation vector, $i_t$ is the input gate's activation vector, $o_t$ is the output gate's activation vector, $h_t$ is the output vector of the LSTM unit and $c_t$ is the cell state vector. $w, u, B $ are the parameters of weight matrices and bias vectors which are learned during the training.

The basic LSTM formulation has been extended to add more complicated LSTM units by adding more contextual information such as social cues (influence of neighboring humans) \cite{gupta2018social, Alahi_2016_CVPR} or environmental cues (influence of scene) \cite{sadeghian2018sophie, xue_wacv2018_sslstm}.  While these have been effective in improving prediction performance, they still utilize the LSTM which has hidden state $h_t$ dependent on previous time-steps and can not be parallelized.  Sequential evaluation limits the speed of any LSTM-based architecture.

\begin{figure}
\centering
\includegraphics[width=0.9\textwidth]{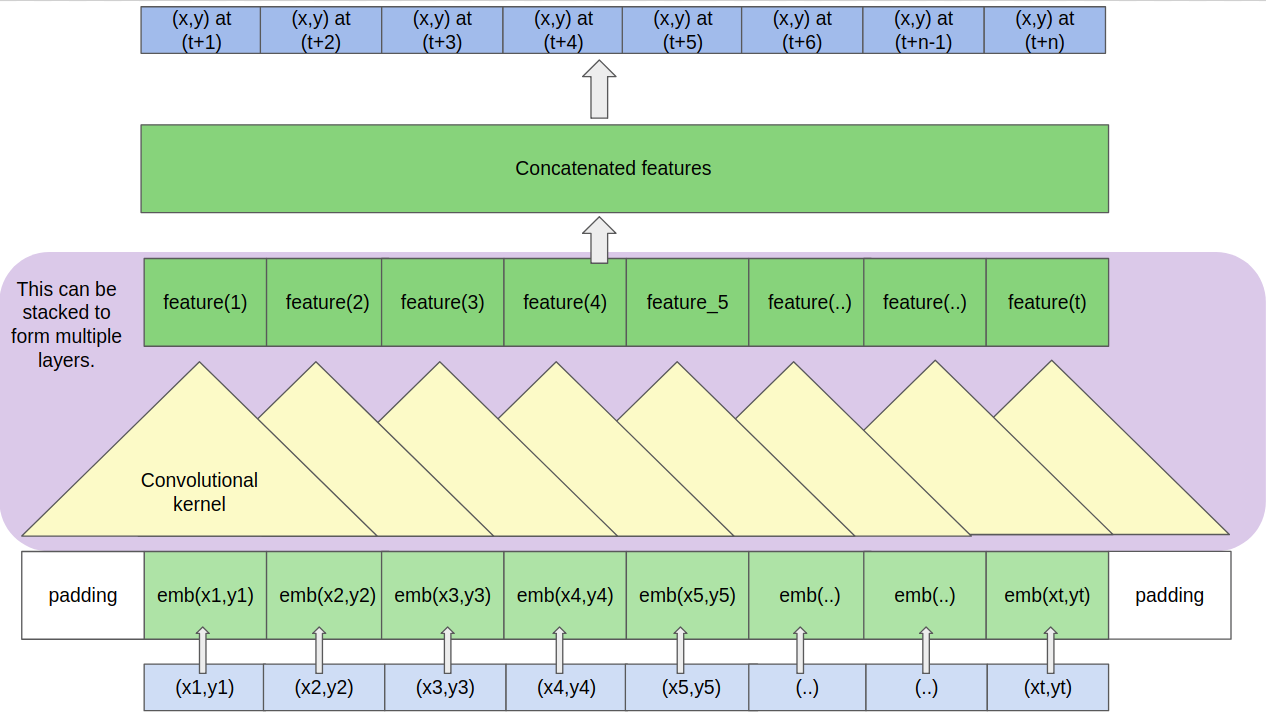}
\caption{Our convolutional model for trajectory prediction. Note that all operations are feed-forward in nature and hence can be parallelized.}
\label{fig:cnn_pred_model}
\end{figure}
\subsection{CNN-Based Framework}

In contrast with LSTM-based networks, our proposed network (Fig. 1) utilizes highly parallelizable convolutional layers to handle temporal dependencies.  The CNN-network is actually a simple sequence-to-sequence architecture.  Trajectory histories are used as input and embedded to a fixed size through a fully connected layer.  Convolutional layers are stacked and used to enforce temporal consistency.  Finally, the features from the final convolutional layer are concatenated and passed through a fully connected layer to generate all predicted positions $\left(x^t, y^t\right)_{t=t+1}^{t+t_{pred}}$ at once.  

The model is inspired by the work of \cite{AnejaConvImgCap17} which has to predict a discrete output for the neural machine translation task.  In that setting, the output at the next time step is highly dependent on the current time step for grammatical coherency and CNNs performed well.  The major differences between this work and theirs are that trajectory prediction provides continuous output rather than discrete items and our architecture predicts all future time steps at once. We constantly pad the input to convolution layer such that output from the convolutional layer is of the same size as input to the layer. This way, we can build a neural network as deep as we want. We build the network deep enough to capture the context from every time step of the observed trajectory. We discuss more in the next subsection.

Through an ablation study (Section \ref{sec:qualitative}), we found that predicting one time step at a time leads to worse results than all future times at once.  We believe this is due to error of the current prediction being propagated forward in time in a highly correlated fashion.  Also, unlike LSTM-based architectures which utilize a recurrent function to compute sequentially, all the computation in the proposed model are feed-forward in nature resulting in a significant performance boost with respect to inference time.  Additionally, the convolutions can be easily parallelized.  


\begin{figure}
\centering
\includegraphics[width=0.9\textwidth]{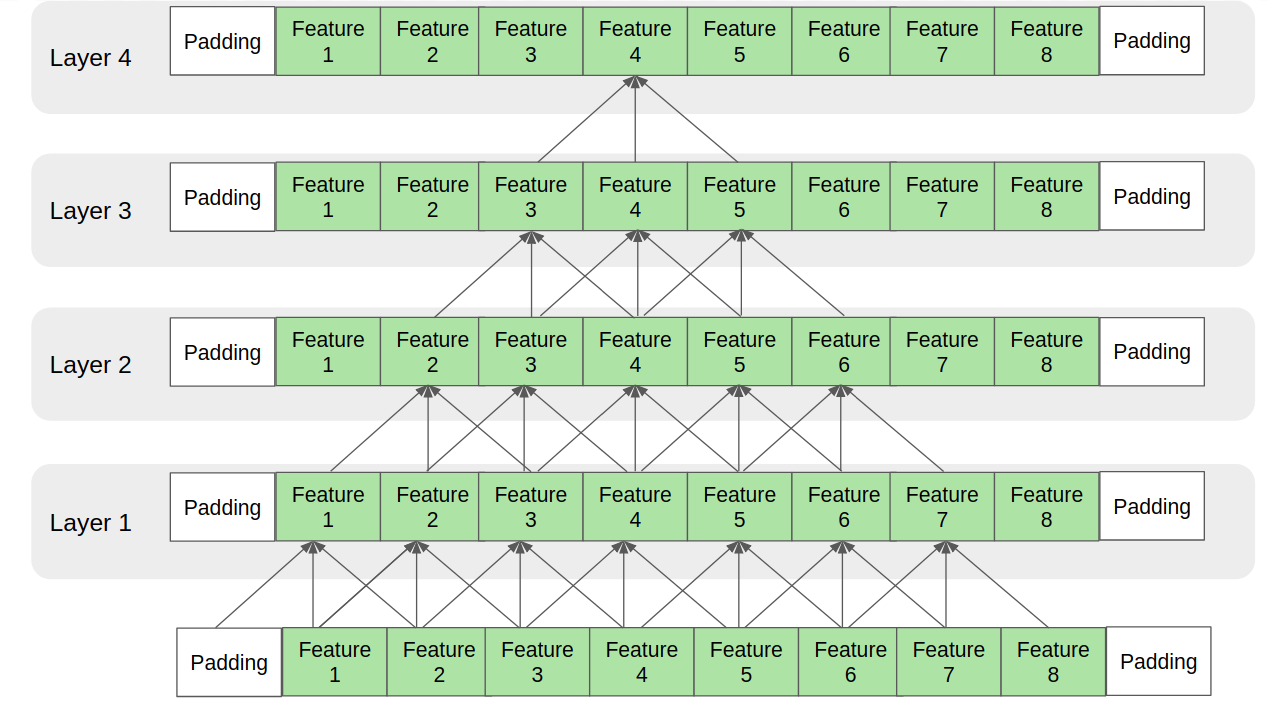}
\caption{For an input having eight temporal dimension and convolutional layers having kernel size of three, we need at least four layer to capture the context from all time-steps.}
\label{fig:cnn_pred_model}
\end{figure}

\subsection{Implementation Details}

We use a kernel size of 3 for all the kernels, by ablation study we found that it works better than other odd kernel sizes when we apply symmetric padding. As the observed trajectory length is eight for all the experiments we conduct, we use a four-layered convolutional network. As shown in Fig. 2, All the features in layer 4 capture context from all eight trajectory observations.  
Unlike temporal convolutional networks (TCNs), we do not use dilated convolutions because we do not want to lose information on such a small temporal dimension. Additionally, we use full rather than causal kernels since the output is a prediction. The embedding layer which converts the geometrical coordinates to embeddings has a dimension of 32, and the subsequent convolution layers produce outputs of the same dimensions. For optimization, we use Adam\cite{journals/corr/KingmaB14} with a learning rate of 0.001. We use a batch size of 32. The model is trained until the validation loss (L2 loss) stops decreasing.

\section{Experiments}

Following common practice in literature (e.g. \cite{gupta2018social}), experimental evaluation is conducted on publically available pedestrian trajectory datasets.  Evaluation utilizes eight historical samples (3.2 s) to give a long-term prediction of the next 12 samples (4.8 s).  

\subsection{Datasets and Evaluation Criteria}
Two publicly available datasets which provide over 1500 pedestrian trajectories in varied crowd settings are utilized in our experiments.  The ETH dataset \cite{Pellegrini:2010:IDA:1886063.1886098} consists of the ETH and HOTEL scenes while the UCY dataset \cite{lealcvpr2014} has the UNIV, ZARA1, and ZARA2 scenes.  The trajectories are rich with challenging human-human interaction scenarios such as group behavior, non-linear trajectories, people crossing paths, collision avoidance, and group formation and dispersion.  All trajectory data has been converted from image to real-world coordinates and interpolated at 2.5 Hz. 


As with prior work \cite{Alahi_2016_CVPR} \cite{Lee2017DESIREDF}, we use two metrics for computing prediction error:

\begin{enumerate}

\item Average Displacement Error (ADE): Computes the mean of euclidean distance between the points in predicted trajectory and the corresponding points in ground truth for all predicted time steps.
$$
\newcommand\norm[1]{\left\lVert#1\right\rVert}
ADE = \frac{\sum_{t = obs + 1}^{T_{pred}} \norm{Y_t - \hat{Y}_t }}{T_{pred} - T_{obs}}
$$

\item Final Displacement Error (FDE): The Euclidean distance between final destination as per the ground truth and the predicted destination at end of the prediction period $T_{pred}$.
$$
\newcommand\norm[1]{\left\lVert#1\right\rVert}
FDE = \norm{Y_{T_{pred}} - \hat{Y}_{T_{pred}}}
$$
\end{enumerate}

Similar to \cite{gupta2018social} \cite{Alahi_2016_CVPR}, we follow leave-one-out approach. We train on four of the five crowd scenes and test on the remaining set. The trajectory is observed for 8-time steps (3.2 seconds), then the model makes the prediction for 12-time steps (4.8 seconds).

\begin{table}
\centering
\caption{Quantitative ADE/FDE for the task of predicting 12 future time steps given 8 previous time steps. More contextual information is provided from left to right ($^+$social, $^{++}$raw scene image)
}
\label{table:1}
\begin{tabular}{|c| c | c | c | c | c | c |} 
 \hline
 \textbf{Dataset} & \textbf{Ours} & \textbf{LSTM} & \textbf{S-GAN$^+$} & \textbf{S-LSTM$^+$} & \textbf{S-GAN-P$^+$}  &\textbf{SoPhie$^{++}$}\\
 \hline\hline
 \textbf{ETH} & 1.04/2.07 & 1.09/2.41 & {0.81/1.52} & 1.09/2.35 & 0.87/1.62 & \textbf{{0.70/1.43}}\\  
 \textbf{HOTEL} & \textbf{0.59/1.17} & 0.86/1.91 & 0.72/1.61 & 0.79/1.76 & 0.67/1.37 & 0.76/1.67\\
 \textbf{UNIV} & {0.57/1.21} & 0.61/1.31 & 0.60/1.26 & 0.67/1.40 & 0.76/1.52 & \textbf{{0.54/1.24}}\\
 \textbf{ZARA1} & 0.43/0.90 & 0.41/0.88 & {0.34/0.69} & 0.47/1.00 & 0.35/0.68 & \textbf{{0.30/0.63}}\\
 \textbf{ZARA2} & \textbf{0.34/0.75} & 0.52/1.11 & 0.42/0.84 & 0.56/1.17 & 0.42/0.84 & 0.38/0.78\\
 \hline
 \textbf{AVG} & 0.59/1.22 & 0.70/1.52 &  {0.58/1.18} & 0.72/1.54 & 0.61/1.21 & \textbf{{0.54/1.15}}\\
 \hline
\end{tabular}
\end{table}
\subsection{Quantitative Evaluation}
\label{sec:quantitative}

In Table 1, we compare prediction results against five different architectures:
\begin{enumerate}

\item LSTM: A simple Long Short-Term Memory architecture without any pooling mechanism, i.e. it doesn't consider any social context.

\item S-LSTM \cite{Alahi_2016_CVPR}: This model combines LSTMs with a social pooling mechanism to provide social context in a fixed rectangular grid.

\item S-GAN \cite{gupta2018social}: This model uses LSTM and variable social max-pooling mechanism in a Generative Adversarial Network (GAN) architecture to generate multiple plausible trajectories.  The S-GAN-P variant also uses a social pooling mechanism.

\item SoPhie \cite{sadeghian2018sophie}: Apart from having LSTM and pooling for features in a GAN setting, this model applies a scene attention mechanism over the features extracted from images of the scene to augment trajectory information.  
\end{enumerate}
The results are organized by increasing contextual information (e.g. social pooling or raw images for scene information) from left to right.  Note: that LSTM, S-LSTM, and S-GAN results in Table \ref{table:1} were reported in \cite{gupta2018social}.  

We find that our model consistently outperforms the LSTM baseline even though they are utilizing the same basic position inputs.  We speculate that this is because the CNNs do a better job at handling long-term dependencies than the LSTM specifically for continuous numerical regression where the notion of state is not complicated.  Interestingly, ours is the best performing architecture for the HOTEL scene even without the use of social or environmental cues.  This is likely due to the simplicity of the scene since even a simple linear regressor provides better results (0.39/0.72) than all reported here \cite{gupta2018social}.  However, the simple CNN still performs very well even in more complicated scenarios (UNIV and ZARA2) and actually beats techniques that utilize social context.  The UNIV result is most surprising since it is the most complicated scene with large crowds of people.  In these situations, social context may not be relevant (Fig. 4(b)).  In fact, the average performance is quite similar to S-GAN (provides many plausible trajectories), better than S-GAN-P (multiple trajectories with social pooling), and competitive with SoPhie even though those techniques use social context and scene image context (in the case of SoPhie).  
\begin{table}
\centering
\caption{Speed comparison with other architectures}
\label{table:2}
\begin{tabular}{|l| c | c | c | c | c |} 
 \hline
  & \textbf{LSTM} & \textbf{S-GAN} & \textbf{S-GAN-P} & \textbf{Ours} \\
 \hline
Time (s) & 0.009 & 0.022 & 0.067 & 0.002 \\
Speed-Up & $7.44\times$ & $3.0\times$ & $1\times$ & $33.5\times$ \\ 
 \hline
\end{tabular}
\end{table}

The main advantage of our proposed CNN prediction architecture is the computational efficiency of convolution operations which can be highly parallelized.  A speed comparison is provided in Table \ref{table:2} which reports inference time in seconds and speed up factor with respect to the baseline S-GAN-P.  The high speed of our method makes it well suited for mobile robot applications which need to make predictions in real-time.


\begin{table}
\centering
\caption{CNN layer ablation study}
\label{table:3}
\begin{tabular}{|l| c | c | c |} 
 \hline
  \textbf{Layers} & \textbf{Three} & \textbf{Four} & \textbf{Five} \\
 \hline
 ADE & 0.60 & 0.58 & 0.70 \\
 FDE & 1.30 & 1.20 & 1.40 \\
 \hline
\end{tabular}
\end{table}
Furthermore, to decide the number of layers we trained our architecture with different numbers of convolutional layers. Table \ref{table:3} indicates that four layers performed the best.  
We believe this happens because three-layered networks are not able to capture context from all time-steps and five-layered networks are over-parametrized.

\begin{figure}
\hfill
\subfigure[]{\includegraphics[width=6cm]{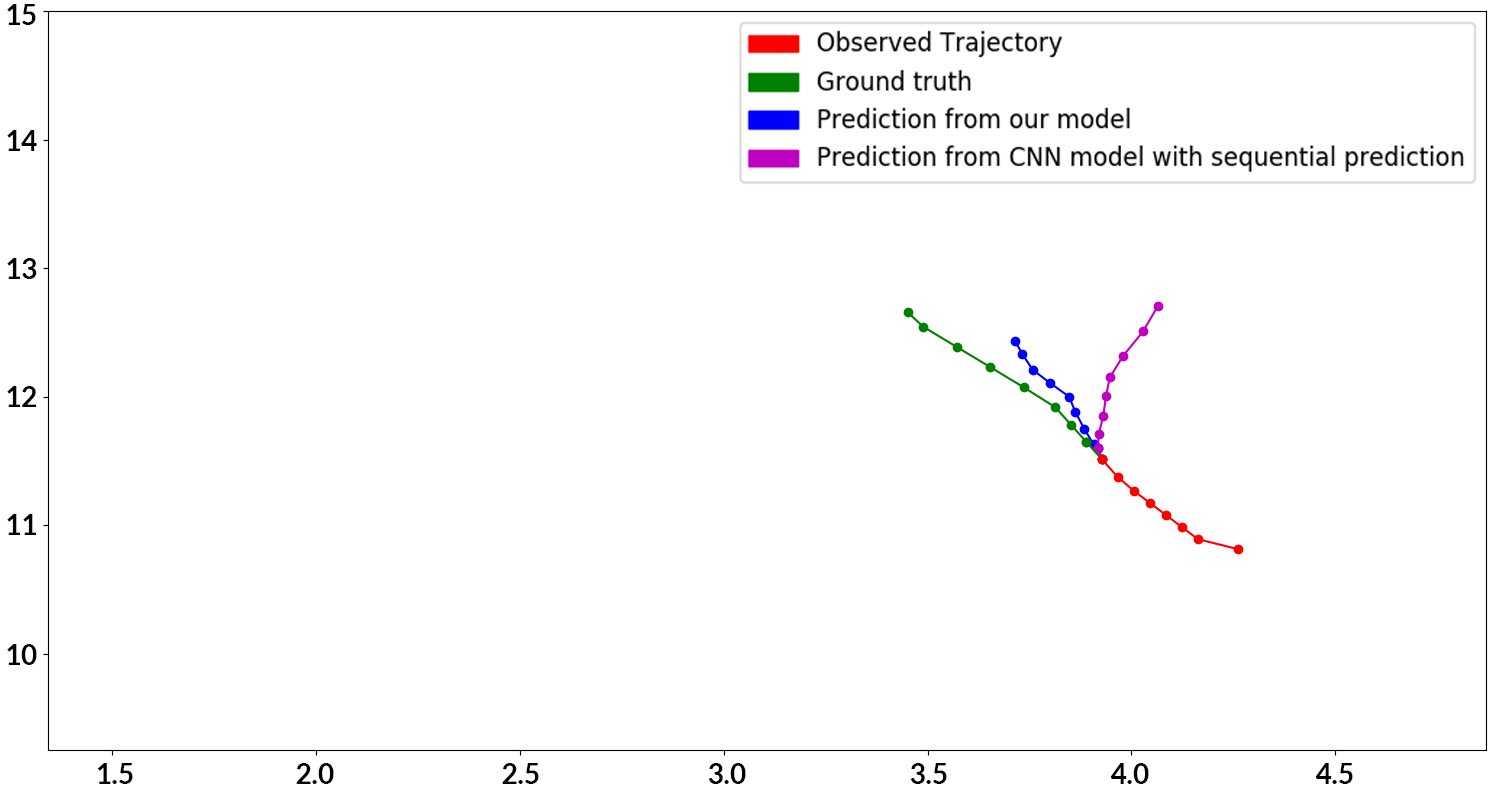}}
\hfill
\subfigure[]{\includegraphics[width=6cm]{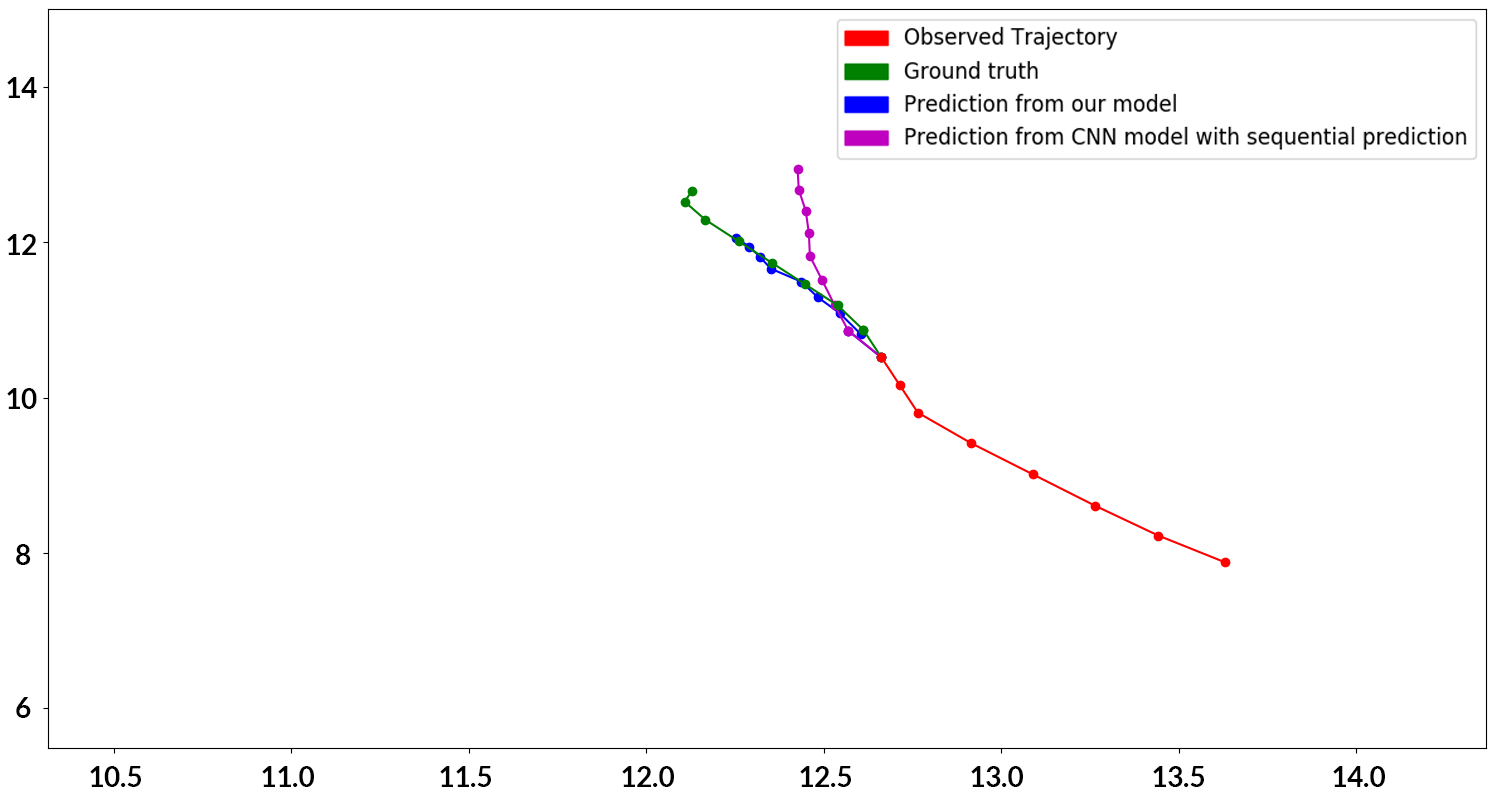}}
\hfill
\hfill
\caption{Multi vs. Sequential Output.  Trajectory prediction sequentially point-by-point performs poorly due to error propogation to future time-steps (trajectory curves off).  Our multi-output model tends to be more resistant to such error accumulation.}
\label{fig:sequential}
\end{figure}

\label{sec:qualitative}
\begin{figure}
\hfill
\subfigure[]{\includegraphics[width=6cm]{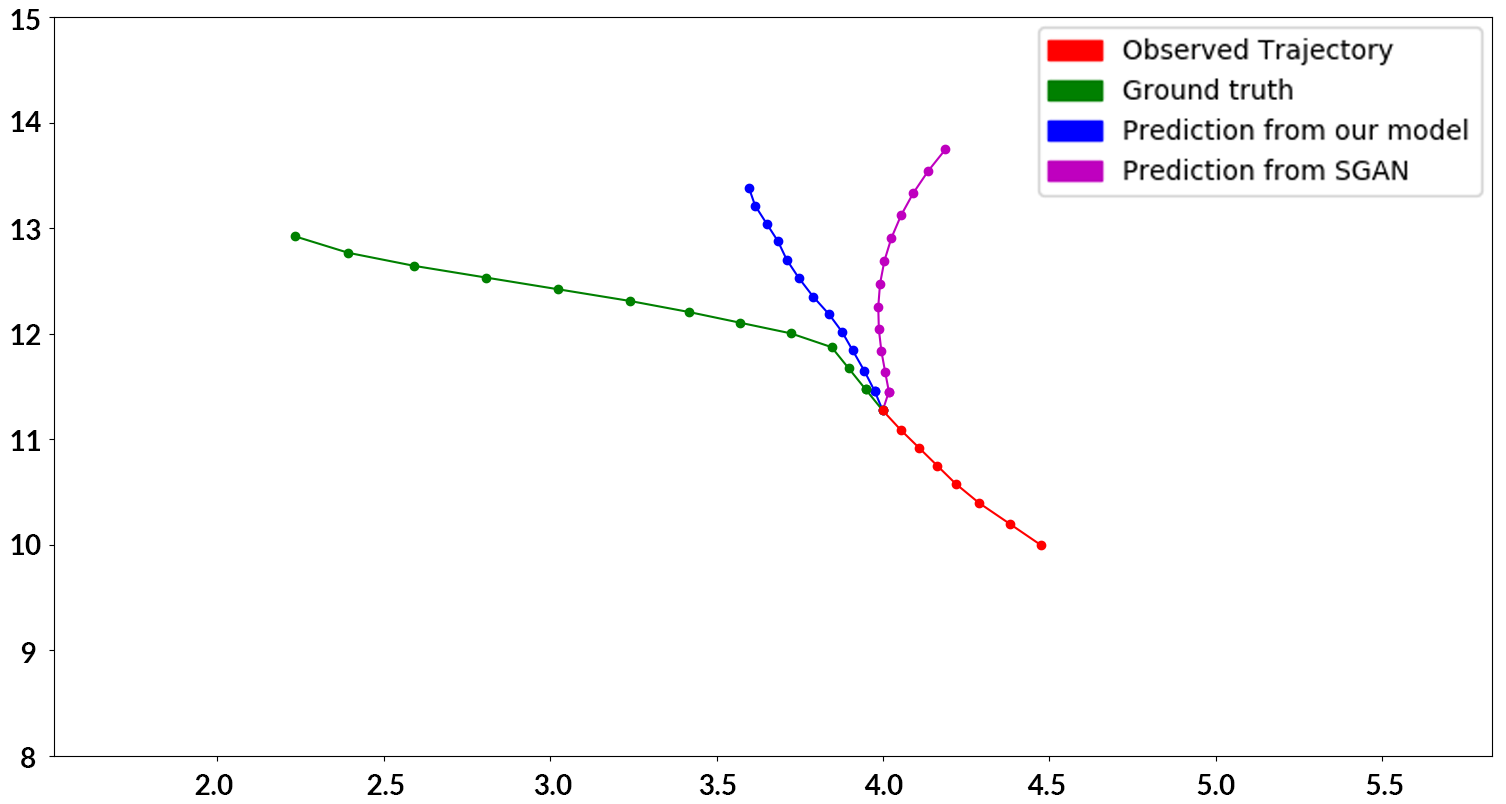}}
\hfill
\subfigure[]{\includegraphics[width=6cm]{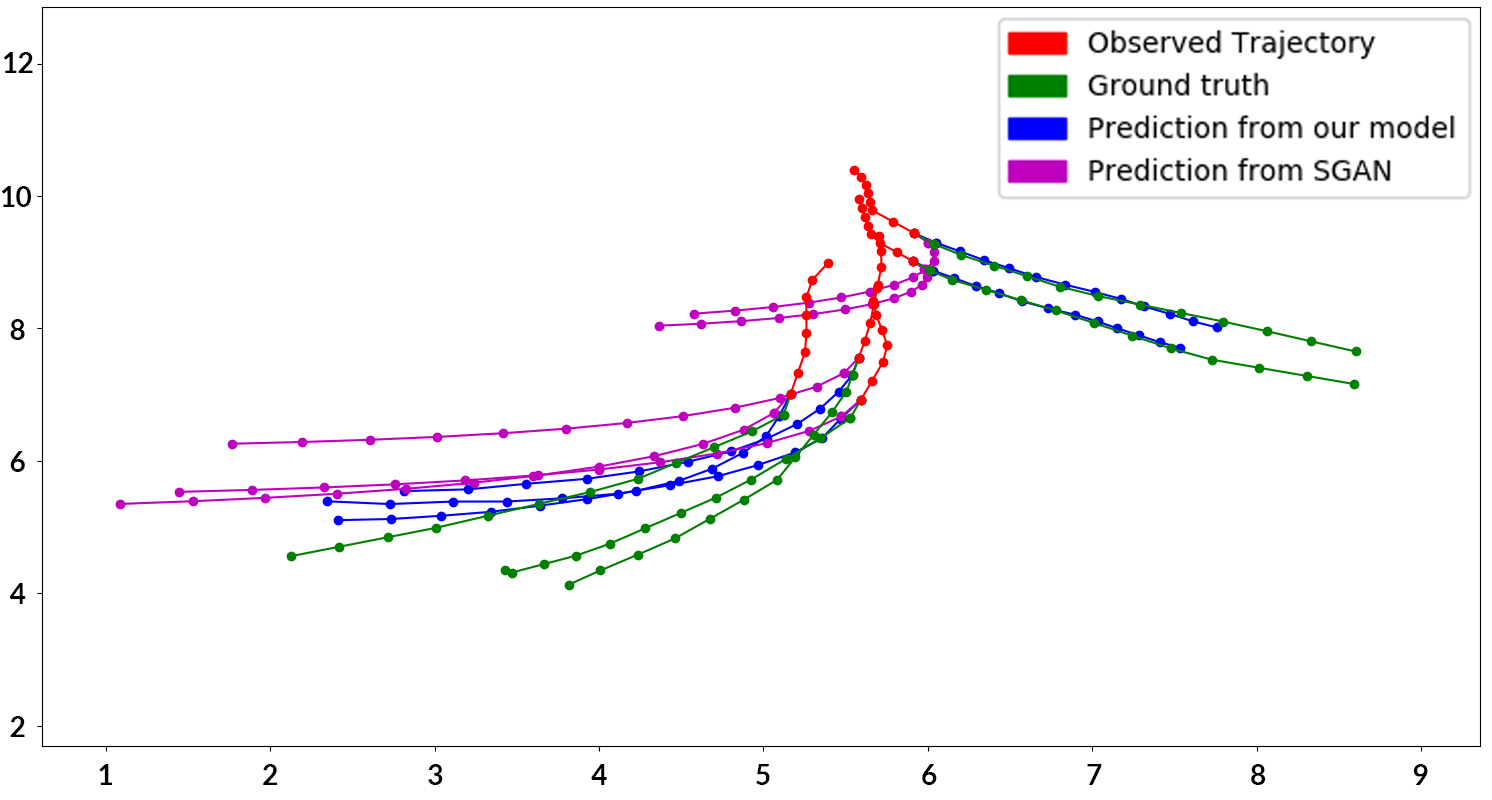}}
\hfill
\subfigure[]{\includegraphics[width=6cm]{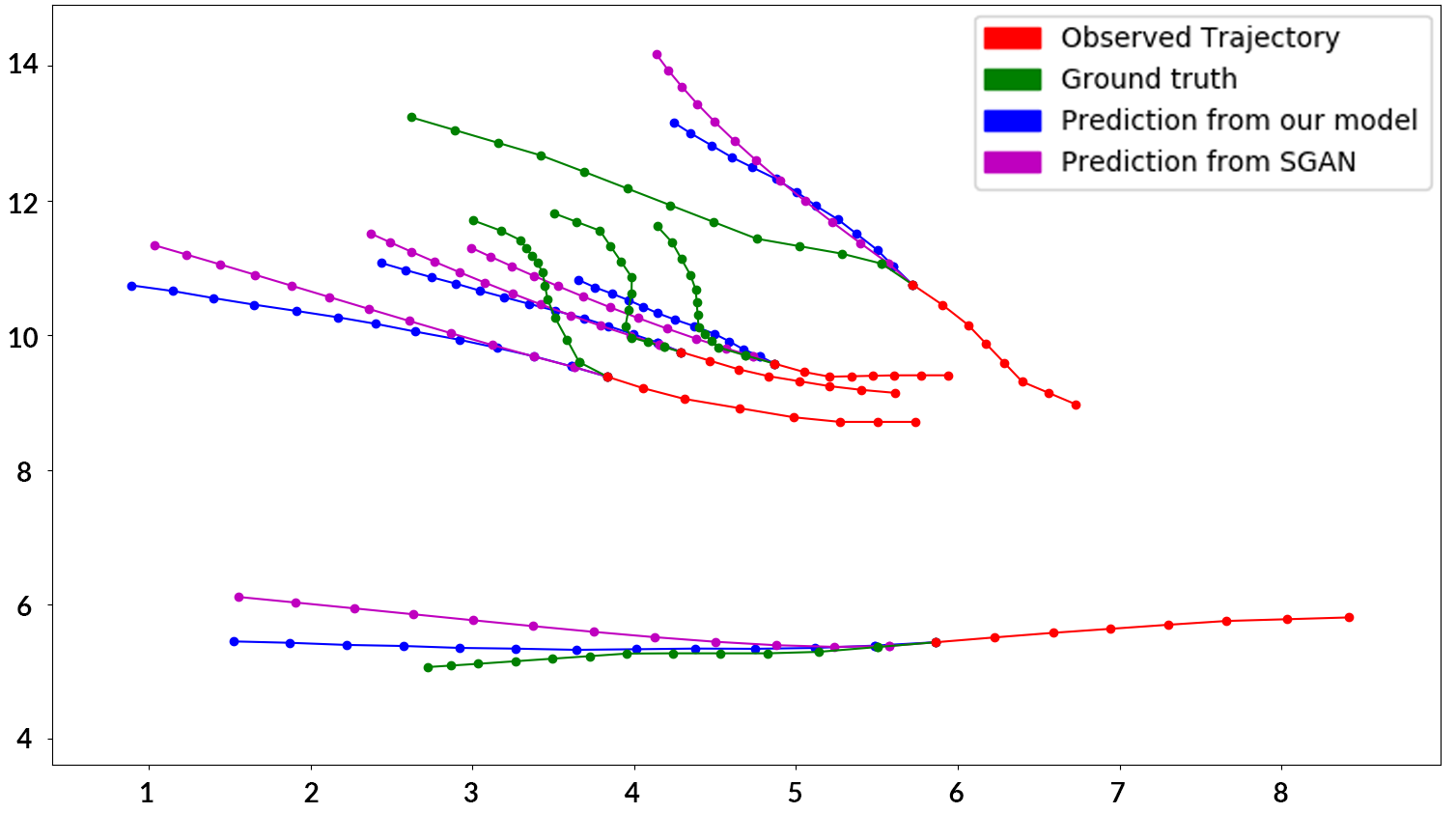}}
\hfill
\subfigure[]{\includegraphics[width=6cm]{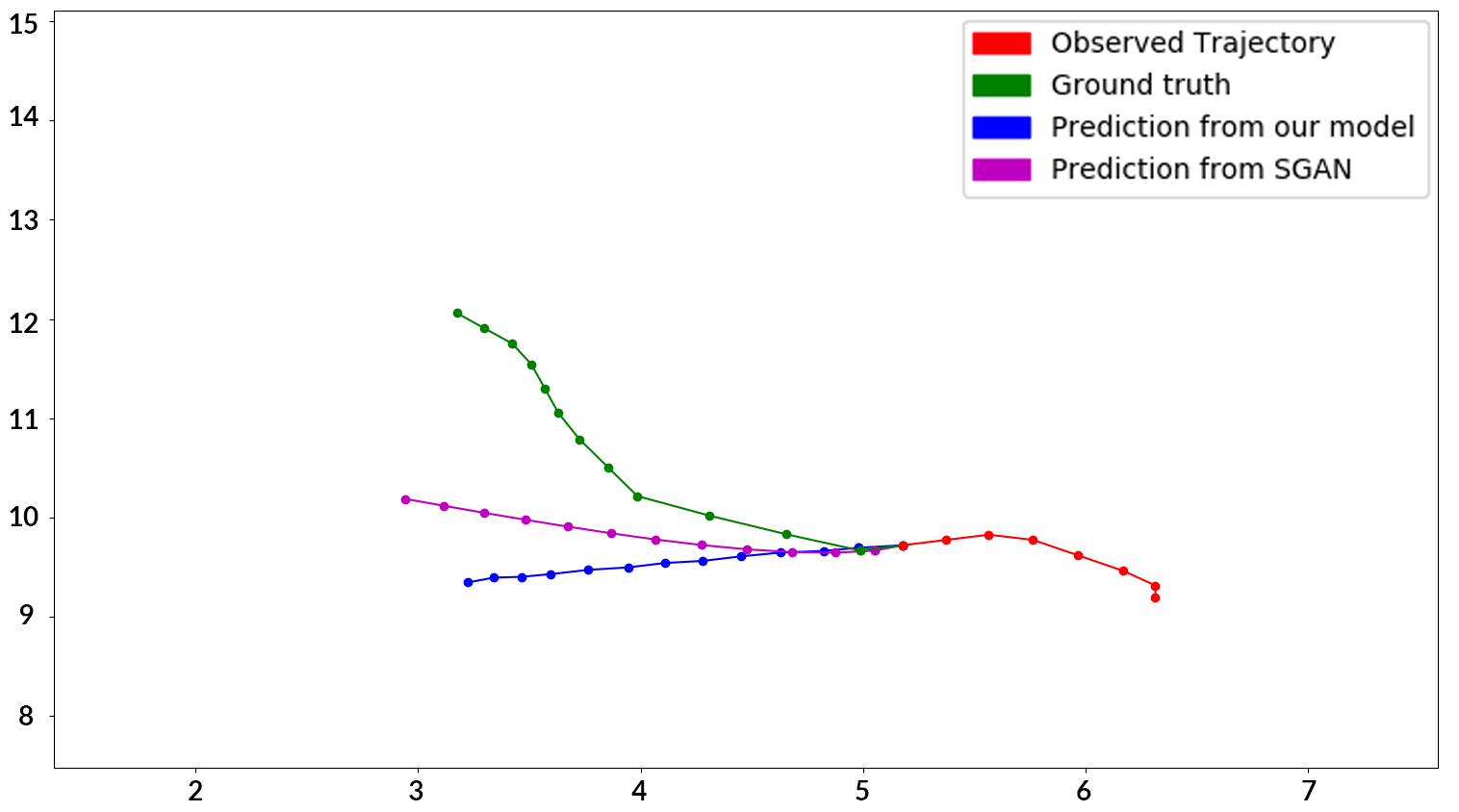}}
\hfill
\caption{Qualitative Comparison on UNIV.  (a) CNN model is better able to interpolate while the S-GAN model seems to accumulate error in subsequent time-steps. (b) Social pooling erroneously combines all five pedestrians and thinks they all should be moving left.  Without pooling, the CNN model is able to better predict the two pedestrians moving right. (c) Both models do a poor job of prediction, especially in the center. (d) SGAN provides a better prediction than CNN.}
\label{fig:qualitative}
\end{figure}

\subsection{Qualitative Evaluation}

In Fig. \ref{fig:sequential} and Fig. \ref{fig:qualitative}, we examine the quality of trajectories produced by the CNN architecture.  One important finding was that sequential prediction (similar to LSTM-based models) performed very poorly (Fig. \ref{fig:sequential}).  Prediction error for the maroon curve was propagated forward resulting in trajectories that ``curved off'' over time.  In contrast, the multi-output CNN architecture was more resistant to this type of error accumulation.  

Fig. \ref{fig:qualitative} provides a comparison between the CNN (blue) and S-GAN (maroon).  (a) provides an exmample when the CNN has a better prediction than S-GAN. In (b), S-GAN's social pooling causes poor prediction since it thinks all five pedestrians should be moving as a group.  Their prediction of the two right moving pedestrians is strongly pulled to the left resulting in large error.  In contrast, the CNN is able to independently predict with better results.  The UNIV scene in particular is quite dense making the pooling operation challenging.  In (c), both CNN and S-GAN fail as seen in the center.  In particular, three pedestrians seem to move in unison to avoid something in the scene and therefore neither algorithm is aware.  Finally, (d) shows an example of S-GAN performing better than CNN.  It is interesting to note that for both (a) and (d), neither technique is actually working that well.  Also, it is difficult to fully understand what is happening without overlaying the trajectories on the image frame.  This strongly hints that trajectories alone (even with social pooling) is not sufficient to make robust prediction. 

Note that unlike state-of-the-art architectures (e.g. S-GAN and SoPhie), our CNN prediction architecture does not include any context outside of individual trajectory information.  Similar social pooling schemes could be added and further improvements are expected.  Additionally, S-GAN reported a $49\times$ speed up over S-LSTM which would make our CNN architecture $500\times$ S-LSTM.

\section{Conclusions}

We present a convolutional architecture based neural network model for trajectory prediction. The simple model gives competitive results with the current state-of-art LSTM-based models while providing better inference time performance. We hope that following this work, more people would be interested in utilizing clever convolutional architectures for trajectory prediction.  Given the current architecture is quite simple, future work will examine the use of dilated convolutions to decrease the number of layers while maintaining the same receptive field and incorporating social context into the model. 

\bibliographystyle{splncs}
\bibliography{egbib}
\end{document}